\pdfoutput=1

\documentclass[11pt]{article}

\usepackage[final]{acl}

\usepackage{times}
\usepackage{latexsym}
\usepackage{amsmath}
\usepackage{url}
\usepackage[most]{tcolorbox}

\usepackage{pbalance}

\usepackage[T1]{fontenc}

\usepackage[utf8]{inputenc}

\usepackage{microtype}

\usepackage{inconsolata}

\usepackage{graphicx}

\title{Prompt-Guided Turn-Taking Prediction}

\author{
 \textbf{Koji Inoue},
 \textbf{Mikey Elmers},
 \textbf{Yahui Fu},
 \textbf{Zi Haur Pang},
 \textbf{Divesh Lala}, \\
 \textbf{Keiko Ochi},
 \textbf{Tatsuya Kawahara}
\\
\\
 Graduate School of Informatics, Kyoto University, Japan, \\
 \small{
   \textbf{Correspondence:} \href{mailto:inoue@sap.ist.i.kyoto-u.ac.jp}{inoue@sap.ist.i.kyoto-u.ac.jp}
 }
}

\begin{document}
\maketitle
\begin{abstract}
Turn-taking prediction models are essential components in spoken dialogue systems and conversational robots.
Recent approaches leverage transformer-based architectures to predict speech activity continuously and in real-time.
In this study, we propose a novel model that enables turn-taking prediction to be dynamically controlled via textual prompts.
This approach allows intuitive and explicit control through instructions such as ``\texttt{faster}'' or ``\texttt{calmer},'' adapting dynamically to conversational partners and contexts.
The proposed model builds upon a transformer-based voice activity projection (VAP) model, incorporating textual prompt embeddings into both channel-wise transformers and a cross-channel transformer.
We evaluated the feasibility of our approach using over 950 hours of human-human spoken dialogue data.
Since textual prompt data for the proposed approach was not available in existing datasets, we utilized a large language model (LLM) to generate synthetic prompt sentences.
Experimental results demonstrated that the proposed model improved prediction accuracy and effectively varied turn-taking timing behaviors according to the textual prompts.
\end{abstract}

\renewcommand{\thefootnote}{\fnsymbol{footnote}}
\footnote[0]{This paper has been accepted for presentation at SIGdial Meeting on Discourse and Dialogue 2025 (SIGDIAL 2025) and represents the author's version of the work.}
\renewcommand{\thefootnote}{\arabic{footnote}}

\section{Introduction}

Turn-taking, which governs transitions between speakers, is fundamental to smooth and natural human communication~\cite{skantze2021review}.
Accurate turn-taking prediction is particularly crucial in developing spoken dialogue systems and conversational robots, as it directly influences interaction quality by minimizing inappropriate interruptions and reducing response delays~\cite{ter2011agents,khouzaimi2015optimising,tisserand2024unraveling,addlesee2025building,skantze2025hri,inoue2025noise}.
Recent advancements, notably transformer-based models such as TurnGPT~\cite{erik2020turngpt} and voice activity projection (VAP)~\cite{erik2022vap}, have significantly improved continuous and real-time speech activity predictions by effectively utilizing conversational history.

In this paper, we introduce a novel transformer-based model capable of dynamically adjusting its turn-taking predictions based on textual prompts.
Our approach integrates prompt embeddings into both channel-wise and cross-channel transformer architectures in VAP, enabling explicit and intuitive control through simple instructions like ``\texttt{faster}'' or ``\texttt{calmer}''.
Previous research indicates that turn-taking behaviors in conversations vary according to individuals' attributes like age, gender, and personality traits such as extraversion and introversion~\cite{Anderson1998,levinson2015timing,su2016exploiting,Liesenfeld2020,Lourenco2023}.
The adaptability of our proposed model thus facilitates such responsive interactions tailored to dialogue contexts, user types, and system configurations.
To our knowledge, this work is the first to explicitly direct turn-taking predictions using textual prompts.

\begin{figure}[t]
    \centering
    \includegraphics[width=\linewidth]{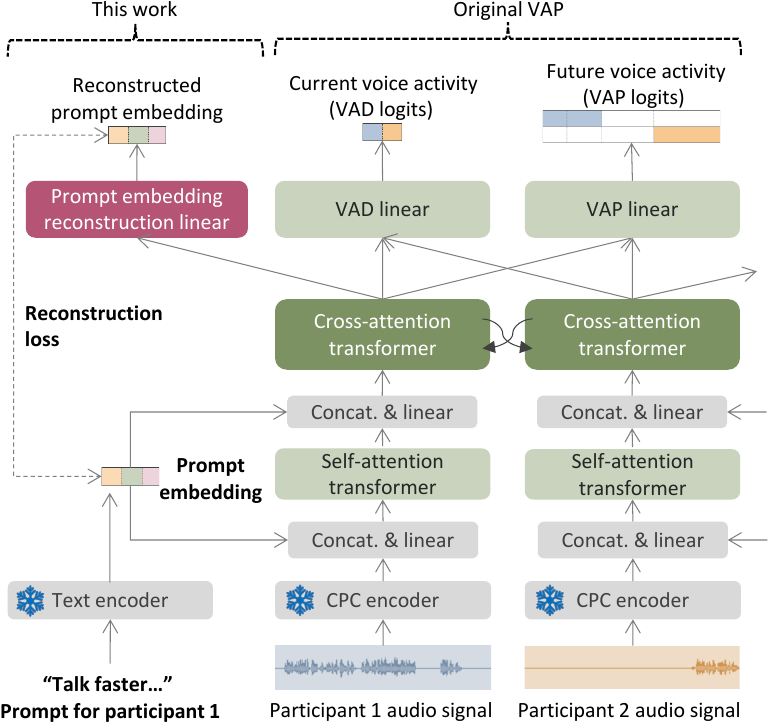}
    \caption{Architecture of the proposed model. The prompt processing (illustrated on the left side) is also applied to those of participant 2.}
    \label{fig:model}
\end{figure}

\section{Proposed Method}

Our proposed model builds upon the VAP architecture that predicts future voice activity for dyadic conversations~\cite{erik2022vap}.
The key innovation of our approach is the integration of textual prompts, allowing for dynamic control over the model's turn-taking behavior, as depicted in Figure~\ref{fig:model}.

\subsection{Base VAP Model Architecture}

The foundation of our model is the VAP architecture that takes a stereo audio waveform as input, where each channel corresponds to one participant, in the manner of full-duplex systems.
Each audio channel is processed by a pre-trained audio encoder using contrastive predictive coding (CPC)~\cite{riviere2020unsupervised}.
The encoder parameters are frozen while training the VAP.
The encoded features for each channel are independently processed by a self-attention transformer which models intra-speaker temporal dependencies.
The outputs from the channel-wise transformers are then fed into a cross-attention transformer.
This component models the interaction between the two participants by allowing each channel's representation to attend to the other.
The final representation from the cross-attention transformer is passed through two separate linear layers to predict:
\begin{itemize}
    \item {\bf VAP logits} represent a probability distribution over 256 discrete states.
    Each state corresponds to predicted voice activity for each speaker across four future time bins (0-200ms, 200-600ms, 600-1200ms, 1200-2000ms).
    This forms the primary VAP objective, framed as a multi-class classification problem.
    \item {\bf VAD (voice activity detection) logits} are the probability of current voice activity for each speaker at the current time frame.
    This serves as an auxiliary task.
\end{itemize}

For the turn-taking prediction, the predicted VAP state probabilities (derived from the VAP logits) are aggregated to represent near-future and longer-term future voice activity:
\begin{itemize}
    \item $p_{now}$: The aggregated probability representing predicted voice activity within the near future (0-600ms, combining the first two bins) for each participant.
    \item $p_{future}$: The aggregated probability representing predicted voice activity in the longer-term future (600-2000ms, combining the last two bins) for each participant.
\end{itemize}

\subsection{Prompt Integration}

To facilitate prompt-based control, we extend the VAP architecture by integrating textual prompt embeddings for each participant independently.
The embeddings encode provided turn-taking behaviors, such as ``\texttt{faster response}'' or ``\texttt{calmer}''.
In this study, we utilized Sarashina-Embedding-v1-1B\footnote{\url{https://huggingface.co/sbintuitions/sarashina-embedding-v1-1b}} that produces 1792-dimensional normalized sentence vectors.
Since the dimensionality of these embeddings varies according to the embedding model (1792 dimensions in our case), we apply a linear transformation to standardize their dimensions to match the input requirements of the transformer (e.g., 256 dimensions).

Inspired by prompt-guided text-to-speech frameworks \cite{guo2023prompttts}, we concatenate the prompt embeddings with the VAP feature representations in every time frame, at two strategic points within the architecture:
\begin{itemize}
    \item \textbf{Post-Audio Encoder:} the prompt embeddings are initially concatenated with the output of the CPC audio encoder.
    A linear projection subsequently adjusts this concatenated vector to match the input dimension of the channel-wise transformer.
    \item \textbf{Post-Self-Attention:} After the channel-wise self-attention transformer, the prompt embeddings are concatenated once more.
    Another linear layer reduces the dimensionality of this concatenated representation to align with the input dimension of the cross-attention transformer.
\end{itemize}
The resulting prompt-conditioned vectors are then processed by the cross-attention transformer as per the original VAP architecture.

Final outputs for VAP and VAD predictions are derived from the representations produced by the cross-attention transformer.
Additionally, the model includes an auxiliary linear head tasked with reconstructing the original prompt embeddings from the participant-specific outputs of the cross-attention layer.
This reconstruction constitutes an additional training objective, ensuring robust propagation of prompt information throughout the network.

The training loss for the proposed model thus comprises three components:
\begin{align}
L = L_{\text{vap}} + L_{\text{vad}} + L_{\text{prompt}} ~,
\end{align}
where $L_{\text{vap}}$ and $L_{\text{vad}}$ correspond to losses from the original VAP framework, and $L_{\text{prompt}}$ represents the mean squared error (MSE) between input and reconstructed prompt embeddings.
Through concatenation of prompt embeddings with both initial audio features and intermediate post-self-attention representations, we comprehensively condition the model's predictive processing--from individual-channel analysis through cross-channel interactions--on the provided behavioral prompts.

\section{Dataset}

This section describes the datasets used in this study and the method for generating textual prompts for the proposed approach.

\subsection{Spoken Dialogue Corpora}

This study utilized the following three types of Japanese spoken dialogue datasets.
The total duration of these datasets combined is approximately 953.5 hours.
The entire collection of these datasets was randomly split at the session level into training, validation, and test sets, with an 8:1:1 ratio.
By incorporating data that includes various dialogue tasks beyond simple chit-chat, we aimed to train a prompt-controllable model adaptable to various situations.

\paragraph{Online Conversation Dataset} This dataset was newly collected for this research.
It comprises free-form chit-chat dialogues conducted by 109 participants using an online meeting tool.
Each participant engaged in multiple sessions lasting approximately 20 to 30 minutes, with the constraint that each pair of participants interacted only once.
In total, the dataset includes 2,166 dialogues, amounting to 803 hours.
Silero VAD\footnote{\url{https://github.com/snakers4/silero-vad}} was used to annotate utterance segments.

\paragraph{Travel Agency Task Dialogue~\cite{inaba2023travel}} This consists of task-oriented dialogues simulating interactions between customers and staff at a travel agency.
These dialogues were also recorded using an online meeting tool.
The full dataset contains 329 dialogues, totaling 115.5 hours.

\paragraph{Human-Robot Dialogue} This consists of dialogues between humans and the android ERICA, collected using the Wizard-of-Oz (WOZ) method~\cite{inoue2025naacl}.
It encompasses various tasks, including attentive listening dialogues where ERICA acts as the listener, job interviews, and first-meeting dialogues.
The total duration for these tasks combined is approximately 35 hours.

\begin{figure*}[t]
    \centering
    \includegraphics[width=\linewidth]{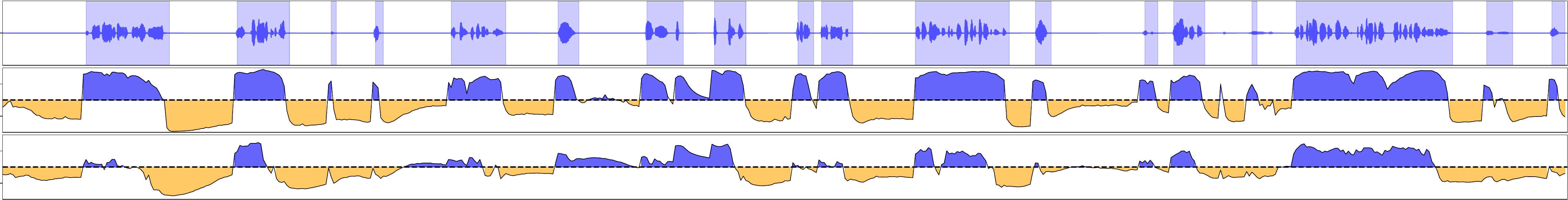}
    \caption{Example output of the proposed method. The top row is a waveform representation of the user's audio input. The middle and bottom rows indicate probabilities for $p_{now}$ and $p_{future}$, respectively, with blue indicating the user and yellow indicating the system.}
    \label{fig:example1}
\end{figure*}

\begin{figure*}[t]
    \centering
    \includegraphics[width=\linewidth]{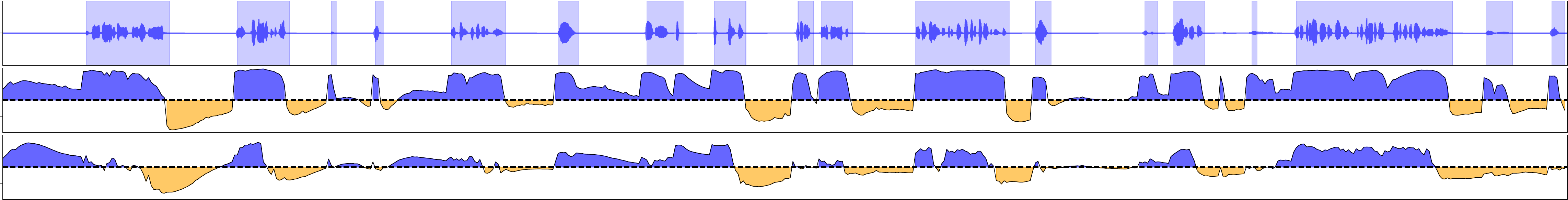}
    \caption{Example output where the input prompts are opposite to those presented in Figure~\ref{fig:example1}.}
    \label{fig:example2}
\end{figure*}

\subsection{Prompt Generation} \label{sec:generation}

Although textual prompts are essential for controlling turn-taking style in our proposed method, such information is typically absent in existing datasets.
Consequently, we synthetically generated prompt sentences from the dialogue data using a large language model (LLM). Specifically, we provided GPT-4.1\footnote{\url{https://platform.openai.com/docs/models/gpt-4.1}} with voice activity details, including speaker identities and utterance timing (start and end times) for each 20-second audio segment aligned with the VAP model's training input format.

Based on this information, we instructed the LLM to first generate a descriptive impression of each participant's turn-taking style using a chain-of-thought methodology (Appendix~\ref{app:prompt}).
Subsequently, the LLM produced a corresponding prompt (instruction) intended to replicate that specific style.
An actual example of the generated output is provided below:
\begin{quote}
\it
\textbf{Impression of participant A}: Speaks less frequently but tends to speak for relatively longer durations once beginning, creating a calm impression with distinct pauses before speaking.

\textbf{Prompt for participant A}: Please speak calmly, with deliberate pauses, delivering substantial content at once.
Include brief silences before and after your utterances.
\end{quote}
The prompt sentences generated using this method were employed for both training and evaluating the proposed model.

\section{Experiment}

We first quantitatively evaluated the performance of the turn-taking prediction by comparing the proposed method against the original VAP model, which does not utilize prompt inputs.
The evaluation metrics included the VAP test loss ($L_{vap}$) and balanced accuracy for predicting turn shifts or holds during mutual silences~\cite{inoue2024coling}. 
Model configurations and training parameters were consistent with those described in the original work~\cite{erik2022vap}.

The results are summarized in Table~\ref{tab:result}.
The proposed method achieved improvements for both VAP loss and balanced accuracy by incorporating prompt information. 
These findings demonstrate that the generated prompt sentences effectively enhanced the accuracy of turn-taking predictions, indicating the architecture's successful integration of additional prompt data.

\begin{table}[t]
    \centering
    \caption{Turn-taking prediction performance. {\it S/H pred.} is balanced accuracy (\%) for turn shift or hold prediction.}
    \begin{tabular}{ccc}
    \hline
    \multicolumn{1}{c}{Method} & VAP loss ($\downarrow$) & S/H pred. ($\uparrow$) \\
    \hline
    Original VAP & 2.431 & 77.17 \\
    Proposed & 2.346 & 79.80 \\
    \hline
    \end{tabular}
    \label{tab:result}
\end{table}

Next, we qualitatively evaluated how the input prompt text influenced the output of the proposed model.
In order to simulate a user-system dialogue and evaluate when the system should take the turn, we used one channel from a test audio as the user's audio input and set the other channel to zero to represent the system's silence.
We prepared two text prompts: one representing the user's instructions or assumed intentions within the dialogue, and the other specifying the system's behavioral setting.

Figure~\ref{fig:example1} illustrates an example output, highlighting when the system should take its turn (yellow area).
In this example, the user prompt is ``{\it Before speaking, take a short pause and think carefully, then begin speaking politely. Do not rush your response; maintain a calm and steady pace.}'' 
Conversely, the system prompt instructs, ``{\it Speak with a good rhythm and respond immediately after the other person finishes speaking. Try to speak more frequently and take the lead in the conversation.}''
The system’s prompt encourages immediate responses and frequent turn-taking, as reflected by the increased frequency and quicker timing of the predicted system turns (yellow).

Figure~\ref{fig:example2} shows another scenario where the prompts between the user and the system were swapped with the same audio input.
With the system now instructed to respond slowly and deliberately, fewer turn shifts were observed compared to the previous example, aligning with the given behavioral prompt.

\section{Conclusion}

We introduced a novel prompt-guided turn-taking prediction model that dynamically adjusts its behavior based on textual prompts integrated into the VAP model.
Our experiments with approximately 950 hours of Japanese dialogue data demonstrated that the model effectively modulates turn-taking behaviors according to different prompts, improving both prediction accuracy and adaptability.

Future work includes integrating this approach into practical dialogue systems and robots to verify its utility, particularly by automatically inferring user prompts or configuring system prompts.
This includes adapting interactions based on user characteristics such as age (child, adult, or elderly) or personality.
Additionally, validating the appropriateness of generated prompts and expanding prompt variations, along with extending the work to include aspects like human alignment and multilingual capabilities~\cite{inoue2024coling}, remain essential next steps.

\section*{Acknowledgments}

This work was supported by JST PRESTO JPMJPR24I4, JST Moonshot R\&D JPMJPS2011, and JSPS KAKENHI JP23K16901.


\section*{Limitations}

A notable limitation of our study is the reliance on synthetically generated textual prompts due to the absence of natural prompts in existing dialogue datasets.
This artificial scenario might limit the generalizability of the model's effectiveness in real-world contexts.
Additionally, practical implementation necessitates incorporating feedback from humans (users or developers) into prompts to effectively reflect real-world usage scenarios and increase the model's controllability.
Furthermore, our current evaluation is primarily based on the sets of Japanese dialogue data, which may restrict the applicability of our findings to other languages and cultural contexts.

\section*{Ethical Considerations}

Our research involves the automated modulation of conversational behaviors, raising ethical considerations around user autonomy and consent.
Implementations of such systems must ensure transparency, allowing users to clearly understand and control how their conversational style and personal data are being interpreted and utilized.
Careful consideration should be given to prevent bias or stereotypes when automatically inferring user characteristics such as age or linguistic background, ensuring equitable and respectful interactions across diverse user groups.


\bibliography{acl_latex}

\begin{thebibliography}{19}
\providecommand{\natexlab}[1]{#1}

\bibitem[{Addlesee and Papaioannou(2025)}]{addlesee2025building}
Angus Addlesee and Ioannis Papaioannou. 2025.
\newblock Building for speech: designing the next-generation of social robots for audio interaction.
\newblock \emph{Frontiers in Robotics and AI}, 11:1--5.

\bibitem[{Anderson and Leaper(1998)}]{Anderson1998}
Kristin~J. Anderson and Campbell Leaper. 1998.
\newblock \href {https://doi.org/10.1023/A:1018802521676} {Meta-analyses of gender effects on conversational interruption: Who, what, when, where, and how}.
\newblock \emph{Sex Roles}, 39(3-4):225--252.

\bibitem[{Ekstedt and Skantze(2020)}]{erik2020turngpt}
Erik Ekstedt and Gabriel Skantze. 2020.
\newblock {T}urn{GPT}: a transformer-based language model for predicting turn-taking in spoken dialog.
\newblock In \emph{International Conference on Empirical Methods in Natural Language Processing (EMNLP) Findings}, pages 2981--2990.

\bibitem[{Ekstedt and Skantze(2022)}]{erik2022vap}
Erik Ekstedt and Gabriel Skantze. 2022.
\newblock {Voice Activity Projection}: {Self-supervised} learning of turn-taking events.
\newblock In \emph{INTERSPEECH}, pages 5190--5194.

\bibitem[{Guo et~al.(2023)Guo, Leng, Wu, Zhao, and Tan}]{guo2023prompttts}
Zhifang Guo, Yichong Leng, Yihan Wu, Sheng Zhao, and Xu~Tan. 2023.
\newblock {PromptTTS}: {C}ontrollable text-to-speech with text descriptions.
\newblock In \emph{International Conference on Acoustics, Speech and Signal Processing (ICASSP)}, pages 1--5.

\bibitem[{Inaba et~al.(2022)Inaba, Chiba, Higashinaka, Komatani, Miyao, and Nagai}]{inaba2023travel}
Michimasa Inaba, Yuya Chiba, Ryuichiro Higashinaka, Kazunori Komatani, Yusuke Miyao, and Takayuki Nagai. 2022.
\newblock Collection and analysis of travel agency task dialogues with age-diverse speakers.
\newblock In \emph{International Conference on Language Resources and Evaluation (LREC)}, pages 5759--5767.

\bibitem[{Inoue et~al.(2024)Inoue, Jiang, Ekstedt, Kawahara, and Skantze}]{inoue2024coling}
Koji Inoue, Bing{'}er Jiang, Erik Ekstedt, Tatsuya Kawahara, and Gabriel Skantze. 2024.
\newblock Multilingual turn-taking prediction using voice activity projection.
\newblock In \emph{Joint International Conference on Computational Linguistics, Language Resources and Evaluation (LREC-COLING)}, pages 11873--11883.

\bibitem[{Inoue et~al.(2025{\natexlab{a}})Inoue, Lala, Skantze, and Kawahara}]{inoue2025naacl}
Koji Inoue, Divesh Lala, Gabriel Skantze, and Tatsuya Kawahara. 2025{\natexlab{a}}.
\newblock {Y}eah, {U}n, {O}h: Continuous and real-time backchannel prediction with fine-tuning of voice activity projection.
\newblock In \emph{North American Chapter of Association for Computational Linguistics (NAACL)}, pages 7171--7181.

\bibitem[{Inoue et~al.(2025{\natexlab{b}})Inoue, Okafuji, Baba, Ohira, Hyodo, and Kawahara}]{inoue2025noise}
Koji Inoue, Yuki Okafuji, Jun Baba, Yoshiki Ohira, Katsuya Hyodo, and Tatsuya Kawahara. 2025{\natexlab{b}}.
\newblock A noise-robust turn-taking system for real-world dialogue robots: {A} field experiment.
\newblock \emph{arXiv preprint, arXiv:2503.06241}.

\bibitem[{Khouzaimi et~al.(2015)Khouzaimi, Laroche, and Lefevre}]{khouzaimi2015optimising}
Hatim Khouzaimi, Romain Laroche, and Fabrice Lefevre. 2015.
\newblock Optimising turn-taking strategies with reinforcement learning.
\newblock In \emph{Annual Meeting of the Special Interest Group on Discourse and Dialogue (SIGDIAL)}, pages 315--324.

\bibitem[{Levinson and Torreira(2015)}]{levinson2015timing}
Stephen~C Levinson and Francisco Torreira. 2015.
\newblock Timing in turn-taking and its implications for processing models of language.
\newblock \emph{Frontiers in psychology}, 6:731.

\bibitem[{Liesenfeld et~al.(2020)Liesenfeld, Parti, Hsu, and Huang}]{Liesenfeld2020}
Andreas Liesenfeld, Gabriella Parti, Yu-Yin Hsu, and Chu-Ren Huang. 2020.
\newblock Predicting gender and age categories in english conversations using lexical, non-lexical, and turn-taking features.
\newblock In \emph{Pacific Asia Conference on Language, Information and Computation (PACLIC)}, pages 22--30.

\bibitem[{Lourenço et~al.(2023)Lourenço, Serra, Coutinho, and Pereira}]{Lourenco2023}
Vânia Lourenço, Joana Serra, Joana Coutinho, and Alfredo~F. Pereira. 2023.
\newblock \href {https://doi.org/10.1016/j.cognition.2023.105568} {Turn-taking in free-play interactions: A cross-sectional study from 3 to 5 years}.
\newblock \emph{Cognition}, 239:105568.

\bibitem[{Riviere et~al.(2020)Riviere, Joulin, Mazar{\'e}, and Dupoux}]{riviere2020unsupervised}
Morgane Riviere, Armand Joulin, Pierre-Emmanuel Mazar{\'e}, and Emmanuel Dupoux. 2020.
\newblock Unsupervised pretraining transfers well across languages.
\newblock In \emph{International Conference on Acoustics, Speech and Signal Processing (ICASSP)}, pages 7414--7418.

\bibitem[{Skantze(2021)}]{skantze2021review}
Gabriel Skantze. 2021.
\newblock Turn-taking in conversational systems and human-robot interaction: {A} review.
\newblock \emph{Computer Speech \& Language}, 67:101178.

\bibitem[{Skantze and Irfan(2025)}]{skantze2025hri}
Gabriel Skantze and Bahar Irfan. 2025.
\newblock Applying general turn-taking models to conversational human-robot interaction.
\newblock In \emph{International Conference on Human-Robot Interaction (HRI)}, pages 859--868.

\bibitem[{Su et~al.(2016)Su, Wu, and Zheng}]{su2016exploiting}
Ming-Hsiang Su, Chung-Hsien Wu, and Yu-Ting Zheng. 2016.
\newblock Exploiting turn-taking temporal evolution for personality trait perception in dyadic conversations.
\newblock \emph{IEEE/ACM Transactions on Audio, Speech, and Language Processing}, 24(4):733--744.

\bibitem[{Ter~Maat et~al.(2011)Ter~Maat, Truong, and Heylen}]{ter2011agents}
Mark Ter~Maat, Khiet~P Truong, and Dirk Heylen. 2011.
\newblock How agents' turn-taking strategies influence impressions and response behaviors.
\newblock \emph{Presence: Teleoperators and Virtual Environments}, 20(5):412--430.

\bibitem[{Tisserand et~al.(2024)Tisserand, Stephenson, Baldauf-Quilliatre, Lefort, and Armetta}]{tisserand2024unraveling}
Lucien Tisserand, Brooke Stephenson, Heike Baldauf-Quilliatre, Mathieu Lefort, and Fr{\'e}d{\'e}ric Armetta. 2024.
\newblock Unraveling the thread: understanding and addressing sequential failures in human-robot interaction.
\newblock \emph{Frontiers in Robotics and AI}, 11:1--19.

\end{thebibliography}

\appendix

\section{Used prompt} \label{app:prompt}

The prompt used to generate the turn-taking behavior prompts described in Section~\ref{sec:generation} is as follows:

\begin{tcolorbox}[
  colback=gray!15,
  colframe=gray!70,
  coltitle=white,
  colbacktitle=gray,
  title=\texttt{Prompt Generation},
  fonttitle=\bfseries\sffamily,
  fontupper=\footnotesize,
  boxrule=0.4mm,
  arc=2mm,
  left=2mm,
  right=2mm,
  top=1mm,
  bottom=1mm,
  sharp corners=south,  
  enhanced jigsaw,
  breakable  %
]

The following is a record, in seconds, of the start and end times of each utterance made by Person A and Person B during a conversation.
Based on the amount of utterance, speaking time, time taken to begin speaking, silence duration, and pacing, the impression of both Person A and Person B will be estimated.
Then, based on those impressions, prompts will be designed for an AI to mimic each person's style of turn-taking.
\\

The output should simply follow the format below, consisting of four lines:

Impression of Person A: <sentence describing the impression of Person A>

Impression of Person B: <sentence describing the impression of Person B>

Prompt for Person A: <prompt sentence for Person A>

Prompt for Person B: <prompt sentence for Person B>
\\

A   ~~~ 10.100  ~~~ 12.510

B   ~~~ 13.813  ~~~ 16.725

A   ~~~ 15.123  ~~~ 15.898

B   ~~~ 17.603  ~~~ 18.245
\\

~~~~~~~ (...)
\\

B   ~~~ 26.135 ~~~ 28.211

A   ~~~ 28.348  ~~~ 29.124

B   ~~~ 29.218  ~~~ 29.875

A   ~~~ 30.158  ~~~ 32.556
\end{tcolorbox}

Note that this is a translated version of the original prompt written in Japanese.
The prompt is followed by dialogue context data, which includes the timing information for utterances of both participants.

\end{document}